\documentclass{article}

\PassOptionsToPackage{numbers, compress}{natbib}

\usepackage[preprint]{neurips_2021}

\usepackage[utf8]{inputenc} 
\usepackage[T1]{fontenc}    
\usepackage{hyperref}       
\usepackage{url}            
\usepackage{booktabs}       
\usepackage{amsfonts}       
\usepackage{nicefrac}       
\usepackage{microtype}      
\usepackage{xcolor}         
\usepackage{amsmath}
\usepackage{amssymb}
\usepackage{subfigure}
\usepackage{graphicx}
\usepackage{tabularx}
\usepackage{multirow}
\usepackage{paralist}
\usepackage{enumitem}
\usepackage{subfloat}

\newcommand{\equspace}{3.0pt}
\newcommand{\fixedvskip}{-3mm}

\newcommand{\ie}{\textit{i.e.}}

\newcommand{\etal}{\textit{et al.}}

\newcommand{\gt}{\textit{gt}~}
\newcommand{\posneg}{\textit{pos/neg}~}

\definecolor{todocolor}{RGB}{0,174,247}
					
\title{A Normalized Gaussian Wasserstein Distance for Tiny Object Detection}

\author{%
  Jinwang Wang \\
  Electronic Information School\\
  Wuhan University\\
  \texttt{jwwangchn@whu.edu.cn} \\
  \And
  Chang Xu \\
  Electronic Information School\\
  Wuhan University\\
  \texttt{xuchangeis@whu.edu.cn} \\
  \And
  Wen Yang\thanks{Corresponding author} \\
  Electronic Information School\\
  Wuhan University\\
  \texttt{yangwen@whu.edu.cn} \\
  \And
  Lei Yu \\
  Electronic Information School\\
  Wuhan University\\
  \texttt{ly.wd@whu.edu.cn} \\
}

\begin{document}

\maketitle

\begin{abstract}
Detecting tiny objects is a very challenging problem since a tiny object only contains a few pixels in size. We demonstrate that state-of-the-art detectors do not produce satisfactory results on tiny objects due to the lack of appearance information. Our key observation is that Intersection over Union (IoU) based metrics such as IoU itself and its extensions are very sensitive to the location deviation of the tiny objects, and drastically deteriorate the detection performance when used in anchor-based detectors. To alleviate this, we propose a new evaluation metric using Wasserstein distance for tiny object detection. Specifically, we first model the bounding boxes as 2D Gaussian distributions and then propose a new metric dubbed Normalized Wasserstein Distance (NWD) to compute the similarity between them by their corresponding Gaussian distributions. The proposed NWD metric can be easily embedded into the assignment, non-maximum suppression, and loss function of any anchor-based detector to replace the commonly used IoU metric. We evaluate our metric on a new dataset for tiny object detection (AI-TOD) in which the average object size is much smaller than existing object detection datasets. Extensive experiments show that, when equipped with NWD metric, our approach yields performance that is 6.7 AP points higher than a standard fine-tuning baseline, and 6.0 AP points higher than state-of-the-art competitors. Codes are available at: \url{https://github.com/jwwangchn/NWD}.

\end{abstract}

\section{Introduction}

Tiny objects are ubiquitous in many real world applications including driving assistance, large-scale surveillance, and maritime rescue. Even though object detection has achieved significant progress due to the development of deep neural networks~\cite{Faster-R-CNN_2015_NIPS,Focal-Loss_2017_ICCV,FCOS_2019_ICCV}, most of them are devoted to detecting objects with normal size. While tiny objects (less than $16\times 16$ pixels in the AI-TOD dataset~\cite{AI-TOD_2020_ICPR}) often exhibit with extremely limited appearance information, which increases difficulty in learning discriminative features, leading to enormous failure cases when detecting tiny objects~\cite{SNIPER_2018_NIPS,AI-TOD_2020_ICPR,TinyPerson_2020_WACV}.

Recent advances for tiny object detection (TOD) mainly focus on improving the feature discrimination~\cite{FPN_2017_CVPR,M2Det_2019_AAAI,DetectoRS_2020_arXiv,PGAN_2017_CVPR,SOD-MTGAN_2018_ECCV,Better_to_Follow_2019_ICCV}. 
Some efforts have been devoted to normalizing the scale of input images to enhance the resolution of small objects and corresponding features~\cite{SNIP_2018_CVPR,SNIPER_2018_NIPS}. While the Generative Adversarial Network (GAN) is proposed to directly generate super-resolved representations for small objects~\cite{PGAN_2017_CVPR,SOD-MTGAN_2018_ECCV,Better_to_Follow_2019_ICCV}. 
Besides, the Feature Pyramid Network (FPN) is proposed to learn multi-scale features to achieve scale-invariant detectors~\cite{FPN_2017_CVPR,M2Det_2019_AAAI,DetectoRS_2020_arXiv}. Indeed, existing approaches have improved TOD performance to some extent, but the precision boost is commonly achieved with additional cost.

 \begin{figure}[t]
    \centering
    \subfigure[Tiny scale object]
    {
        \label{fig:tiny_analysis1}
        \includegraphics[width=0.48\linewidth]{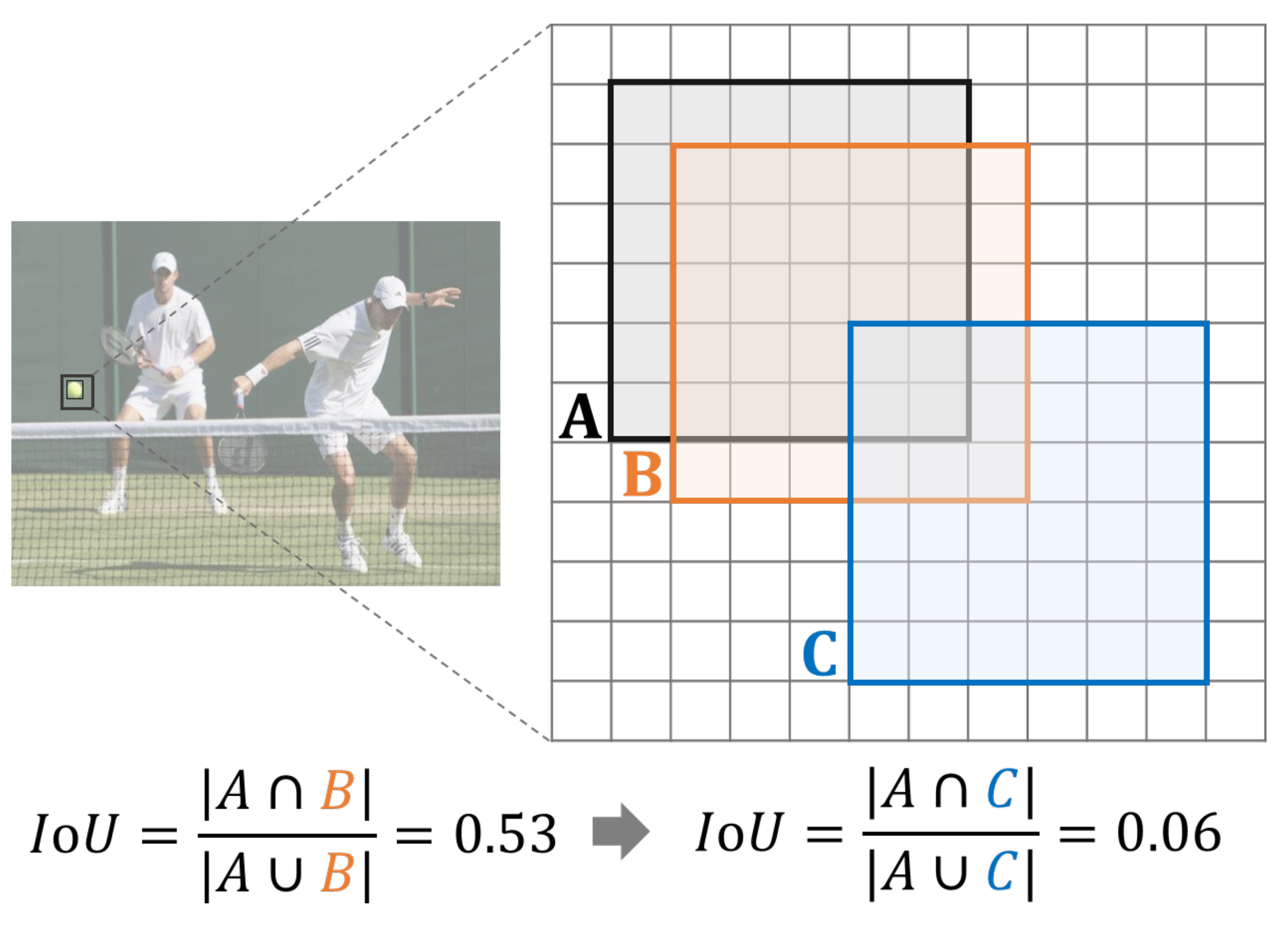}
    }
    \subfigure[Normal scale object]
    {
        \label{fig:tiny_analysis2}
        \includegraphics[width=0.48\linewidth]{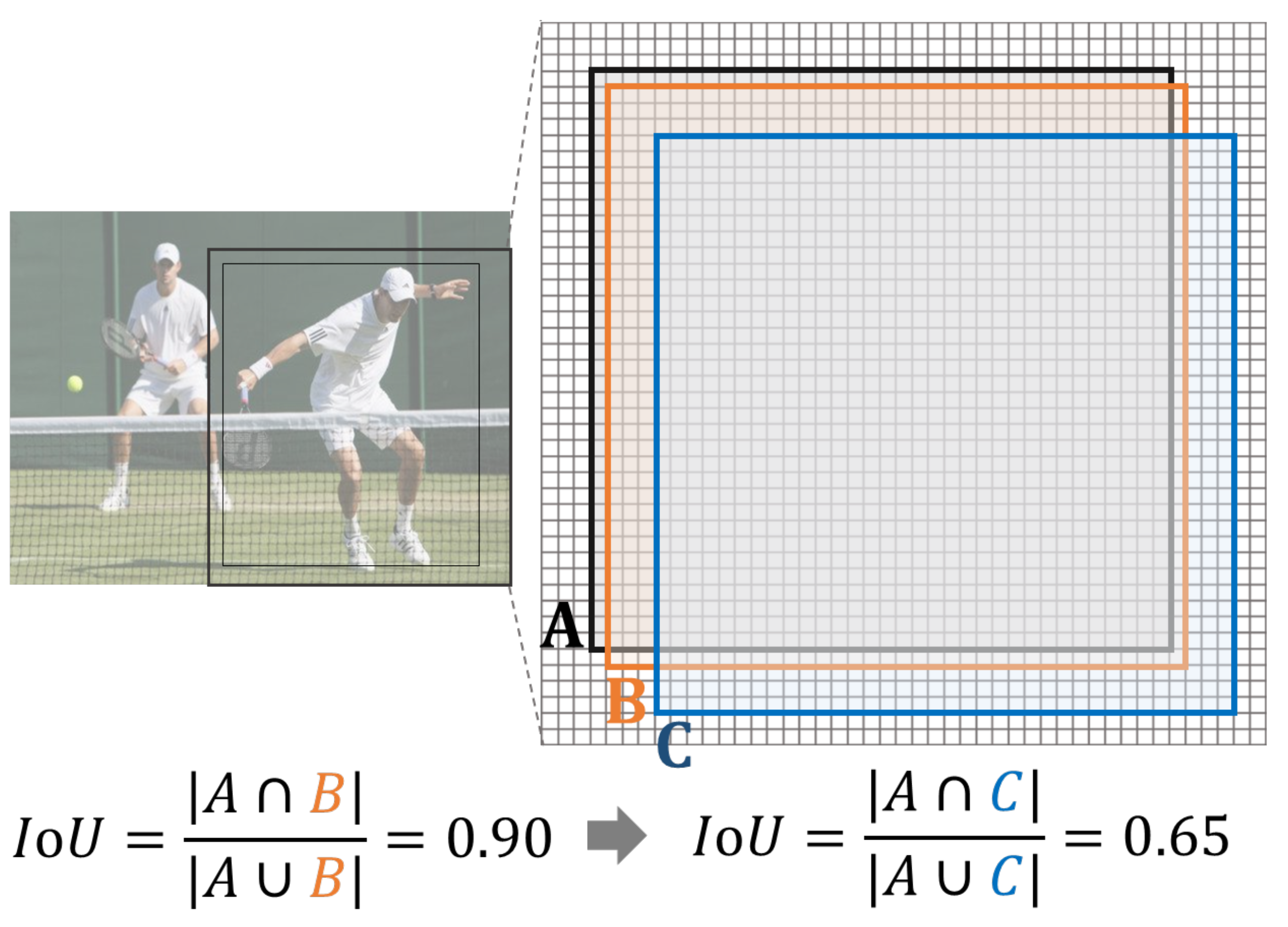}
    }
    \vspace{\fixedvskip}
    \label{fig:tiny_analysis}
    \caption{The sensitivity analysis of IoU on tiny and normal scale objects. Note that each grid denotes a pixel, box $A$ denotes the ground truth bounding box, box $B$, $C$ denote the predicted bounding box with 1 pixel and 4 pixels diagonal deviation respectively.}
    \vspace{\fixedvskip}
\end{figure}

In addition to learning discriminative features, the quality of the training sample selection plays an important role for anchor-based tiny object detectors \cite{atss_2020_cvpr} where the assignment of positive and negative (\textit{pos/neg}) labels is essential. However, for tiny object, the properties of few pixels will increase the difficulty of training sample selection. As shown in Fig.~\ref{fig:tiny_analysis}, we can observe that the sensitivity of IoU to objects with different scales is of great variance. Specifically, for the tiny object with $6\times 6$ pixels, a minor location deviation will lead to notable IoU drop (from $0.53$ to $0.06$), resulting in inaccurate label assignment. However, for the normal object with $36\times 36$ pixels, the IoU changes slightly (from $0.90$ to $0.65$) with the same location deviation. In addition, Fig.~\ref{fig::deviation_analysis} shows four IoU-Deviation curves with different object scales, the curve declines faster as the object size becomes smaller. It is worth noting that, the sensitivity of IoU results from the particularity that the location of bounding box can only change discretely.

This phenomenon implies that IoU metric is no longer invariant to object scale with discretized location deviations and finally leads to the following two flaws in label assignment. Specifically, IoU thresholds ($\theta_p$, $\theta_n$) are used to assign \textit{pos/neg} training samples in anchor-based detectors, and ($0.7$, $0.3$) are used in Region Proposal Network (RPN)~\cite{Fast-R-CNN_2015_ICCV}. Fisrtly, the sensitivity of IoU on tiny object makes a minor location deviation flip the anchor label, leading to \textit{pos/neg} sample features' similarity and the network's difficulty in convergence. Secondly, we find that the average number of positive samples assigned to each ground-truth (\textit{gt}) in AI-TOD dataset~\cite{AI-TOD_2020_ICPR} is less than one using IoU metric since the IoU between some \textit{gt} and any anchor is lower than minimum positive threshold. Therefore, there will be insufficient supervision information for training tiny object detectors. Although dynamic assignment strategies such as ATSS~\cite{atss_2020_cvpr} can adaptively attain IoU thresholds for assigning \textit{pos/neg} labels according to the statistical characteristics of objects, 
the sensitivity of IoU makes it difficult to find a good threshold and provide high-quality \textit{pos/neg} samples for tiny object detectors. 

Observing that IoU is not a good metric for tiny objects, in this paper, we propose a new metric to measure the similarity of bounding boxes by Wasserstein distance to replace standard IoU. Specifically, we firstly model the bounding boxes as 2-D Gaussian distributions, and then use our proposed Normalized Wasserstein Distance (NWD) to measure the similarity of derived Gaussian distributions. The major advantage of Wasserstein distance is that it can measure the distribution similarity even if there is no overlap or the overlap is negligible. In addition, the NWD is insensitive to objects with different scales and thus more suitable for measuring the similarity between tiny objects.

NWD can be applied to both single-stage and multi-stage anchor-based detectors. Besides, NWD can not only replace IoU in label assignment, but also replace IoU in Non-maximum Suppression (NMS) and regression loss function. Extensive experiments on a new TOD dataset AI-TOD~\cite{AI-TOD_2020_ICPR} demonstrate that our proposed NWD can consistently improve the detection performance for all the detectors experimented. The contributions of this paper are summarized as follows.
\begin{itemize}[leftmargin=*]
    \item We analyze the sensitivity of IoU to location deviations of tiny objects, and propose NWD as a better metric for measuring the similarity between two bounding boxes.
    \item We design a powerful tiny object detector by applying NWD to label assignment, NMS and loss function in anchor-based detectors.
    \item Our proposed NWD can significantly improve TOD performance of the popular anchor-based detectors, and it achieves performance improvement from $11.1\%$ to $17.6\%$ on Faster R-CNN on AI-TOD dataset.
\end{itemize}

\section{Related Work}
\label{sec::related_work}

\subsection{Tiny Object Detection}

Most of the previous small/tiny object detection methods can be roughly divided into three categories: multi-scale feature learning, designing better training strategy and GAN-based detection~\cite{reviewtod_2020_icv}.

\textbf{Multi-scale Feature Learning:} A simple and classic way is to resize input images into different scales and to train different detectors, each of which can achieve best performance in a certain range of scales. To reduce the computation cost, some works~\cite{SSD_2016_ECCV, FPN_2017_CVPR, M2Det_2019_AAAI} try to construct feature-level pyramid of different scales. For instance, SSD~\cite{SSD_2016_ECCV} detects objects from feature maps of different resolutions. Feature Pyramid Network (FPN)~\cite{FPN_2017_CVPR} constructs a top-down structure with lateral connections to combine feature information of different scales for improving object detection performance. After that, lots of methods are proposed to further improve FPN performance, including PANet~\cite{PANet_2018_CVPR}, BiFPN~\cite{Efficientdet_2020_CVPR}, Recursive-FPN~\cite{DetectoRS_2020_arXiv}. Besides, TridentNet~\cite{Trident-Net_2019_ICCV} constructs a parallel multi-branch architecture with different receptive fields to generate scale-specific feature maps.

\textbf{Designing Better Training Strategy:} Inspired by the observation that it is difficult to detect tiny objects and large objects simultaneously, Singh~\etal~propose SNIP~\cite{SNIP_2018_CVPR} and SNIPER~\cite{SNIPER_2018_NIPS} to selectively train objects within a certain scale range. Besides, Kim~\etal~\cite{san_2018_eccv} introduce Scale-Aware Network (SAN) and map the features extracted from different spaces onto a scale-invariant subspace, making detectors more robust to scale variation.

\textbf{GAN-based Detectors:} Perceptual GAN~\cite{PGAN_2017_CVPR} is the first to attempt to apply GAN to small object detection, it improves small object detection through narrowing representation difference of small objects from the large ones. Besides, Bai~\etal~\cite{SOD-MTGAN_2018_ECCV}~propose a MT-GAN to train the image-level super-resolution model for enhancing the features of small RoIs. Furthermore, the work in~\cite{Better_to_Follow_2019_ICCV} proposes a feature-level super-resolution approach to improve small object detection performance for proposal based detectors.

\subsection{Evaluation Metric in Object Detection}

IoU is the mostly widely used metric for measuring the similarity between bounding boxes. However, IoU can only work when the bounding boxes have overlap. To handle this problem, generalized IoU (GIoU)~\cite{GIoU_loss_2019_CVPR} is proposed by adding a penalty term of the smallest box converting bounding boxes. Nevertheless, GIoU will degrade to IoU when one bounding box contains another. Thus, DIoU~\cite{diou_2020_aaai} and CIoU~\cite{diou_2020_aaai} are proposed to overcome the limitations of IoU and GIoU by taking three geometric properties into account, \ie, overlap area, central point distance and aspect ratio. GIoU, CIoU and DIoU are mainly applied in NMS and loss function to replace IoU for improving general object detection performance, but the application in label assignment is rarely discussed. In co-current work, Yang~\etal~\cite{gwd_2021_icml} also propose a Gaussian Wasserstein Distance (GWD) loss for oriented object detection by measuring the positional relationship of oriented bounding boxes. However, the motivation of GWD is to solve the boundary discontinuity and square-like problem in oriented object detection. Our motivation is to alleviate the sensitivity of IoU for location deviations of tiny objects and our proposed method can replace IoU in all parts of anchor-based object detectors.

\subsection{Label Assignment Strategies}
It is a challenging task to assign high-quality anchors to \gt boxes of tiny objects. A simple way is to lower the IoU threshold when selecting positive samples. Although it can make tiny objects match more anchors, the overall quality of training samples will deteriorate. Besides, many recent works try to make the label assignment process more adaptive, aiming to improve the detection performance~\cite{ota_2021_arxiv}. For instance, Zhang \etal~\cite{atss_2020_cvpr} propose an Adaptive
Training Sample Selection (ATSS) to automatically compute the \posneg threshold for each \gt by statistic value of IoU from a set of anchors. Kang \etal~\cite{paassignment_2020_eccv} introduce Probabilistic Anchor Assignment (PAA) by assuming that the distribution of joint loss for \posneg samples follows the Gaussian distribution. In addition, Optimal Transport Assignment (OTA)~\cite{ota_2021_arxiv} formulates the label assignment process as an Optimal Transport problem from a global perspective. However, these methods all use IoU metric to measure the similarity between two bounding boxes, and mainly focus on the threshold setting in the label assignment which are not suitable for TOD. In contrast, our method mainly focuses on designing a better evaluation metric which can be used to replace IoU metric in tiny object detectors.

\section{Methodology}

Inspired by the fact that IoU is actually the Jaccard similarity coefficient for computing similarity of two limited sample sets, we design a better metric for tiny objects based on Wasserstein Distance since it can consistently reflect the distance between distributions even if they have no overlap. Therefore, the new metric has better properties than IoU in measuring similarity between tiny objects. The details are as follows.

\subsection{Gaussian Distribution Modeling for Bounding Box}
\label{sec::gaussianmodeling}
For tiny objects, there tend to be some background pixels in their bounding boxes since most~\textit{real objects} are not strict rectangles. In these bounding boxes, foreground pixels and background pixels are concentrated on the center and boundary of the bounding boxes, respectively~\cite{CenterMap-Net_2020_TGRS}. To better describe the weights of different pixels in bounding boxes, the bounding box can be modeled into two dimension (2D) Gaussian distribution, where the center pixel of bounding box has the highest weight and importance of the pixel decreases from the center to the boundary. Specifically, for horizontal bounding box $R=(cx,cy,w,h)$, where $(cx, cy)$, $w$ and $h$ denote the center coordinates, width and height, respectively. The equation of its inscribed ellipse can be represented as 
\begin{equation}
    \setlength{\abovedisplayskip}{\equspace}
	\setlength{\belowdisplayskip}{\equspace}
    \frac{(x-\mu _{x})^2}{\sigma _{x}^{2}}+ \frac{(y-\mu _{y})^2}{\sigma _{y}^{2}}=1,
    \label{eq3.1}
\end{equation}
where $(\mu _{x}, \mu _{y})$ is the center coordinates of the ellipse, $\sigma _{x}$, $\sigma _{y}$ are the lengths of semi-axises along $x$ and $y$ axises. Accordingly, $\mu _{x}=cx, \mu _{y}=cy, \sigma _{x}=\frac{w}{2}, \sigma _{y}=\frac{h}{2}$.

The probability density function of a 2D Gaussian distribution is given by:
\begin{equation}
    \setlength{\abovedisplayskip}{\equspace}
	\setlength{\belowdisplayskip}{\equspace}
   f(\mathbf{x}|\boldsymbol{\mu}, \boldsymbol{\Sigma}) = \frac{ \exp (-\frac{1}{2} (\mathbf{x} - \boldsymbol{\mu})^\intercal \boldsymbol{\Sigma}^{-1} (\mathbf{x} - \boldsymbol{\mu}))}{ 2\pi {|\boldsymbol{\Sigma}|}^{\frac{1}{2}}},
   \label{eq3.2}
\end{equation}
where $\mathbf{x}$, $\boldsymbol{\mu}$ and $\mathbf{\Sigma}$ denote the coordinate $(x, y)$, the mean vector and the co-variance matrix of Gaussian distribution. When
\begin{equation}
    \setlength{\abovedisplayskip}{\equspace}
	\setlength{\belowdisplayskip}{\equspace}
    (\mathbf{x}-\boldsymbol{\mu})^\intercal \mathbf{\Sigma} ^{-1}(\mathbf{x}-\boldsymbol{\mu})=1,
    \label{eq3.4}
\end{equation}
the ellipse in Eq.~\ref{eq3.1} will be a density contour of the 2D Gaussian distribution. Therefore, the horizontal bounding box $R=(cx,cy,w,h)$ can be modeled into a 2D Gaussian distribution $\mathcal{N}(\boldsymbol{\mu},\boldsymbol{\Sigma})$ with 
\begin{equation}
    \setlength{\abovedisplayskip}{\equspace}
	\setlength{\belowdisplayskip}{\equspace}
    \boldsymbol{\mu}=\begin{bmatrix}
        c_x\\ c_y
    \end{bmatrix}
    ,
    \mathbf{\Sigma }=\begin{bmatrix}
    \frac{w^2}{4} & 0 \\ 
    0 & \frac{h^2}{4}
    \end{bmatrix}.
    \label{eq3.3}
\end{equation}

Furthermore, the similarity between bounding box $A$ and $B$ can be converted to the distribution distance between two Gaussian distributions.

\subsection{Normalized Gaussian Wasserstein Distance}
\label{sec::nwd_definition}

We use the Wasserstein distance which comes from Optimal Transport theory to compute distribution distance. For two 2D Gaussian distributions $\mu_{1}=\mathcal{N}(\boldsymbol{m}_{1}, \boldsymbol{\Sigma}_{1})$ and $\mu_{2}=\mathcal{N}(\boldsymbol{m}_{2}, \boldsymbol{\Sigma}_{2})$, the $2^{\rm nd}$ order Wasserstein distance between $\mu_{1}$ and $\mu_{2}$ is defined as:
\begin{equation}
    \setlength{\abovedisplayskip}{\equspace}
	\setlength{\belowdisplayskip}{\equspace}
    W_{2}^{2}(\mu_1, \mu_2)=\left\|\mathbf{m}_{1}-\mathbf{m}_{2}\right\|_{2}^{2}+\textbf{Tr} \left(\mathbf{\Sigma}_{1}+\mathbf{\Sigma}_{2}-2\left(\mathbf{\Sigma}_{2}^{1 / 2} \mathbf{\Sigma}_{1} \mathbf{\Sigma}_{2}^{1 / 2}\right)^{1 / 2}\right),
    \label{eq3.8}
\end{equation}

and it can be simplified as:

\begin{equation}
    \setlength{\abovedisplayskip}{\equspace}
	\setlength{\belowdisplayskip}{\equspace}
    W_{2}^{2}(\mu_1, \mu_2) =\left\|\mathbf{m}_{1}-\mathbf{m}_{2}\right\|_{2}^{2}+\left\|\boldsymbol{\Sigma}_{1}^{1 / 2}-\boldsymbol{\Sigma}_{2}^{1 / 2}\right\|_{F}^{2}, \\
    \label{sec:simplified_w2}
    \
\end{equation}
where $\left \| \cdot  \right \|_{F}$ is the Frobenius norm. 

Furthermore, for Gaussian distributions $\mathcal{N}_a$ and $\mathcal{N}_b$ which are modeled from bounding boxes $A=({cx}_a, {cy}_a, w_a, h_a)$ and $B=({cx}_b, {cy}_b, w_b, h_b)$, Eq.~\ref{sec:simplified_w2} can be further simplified as:
\begin{equation}
    \setlength{\abovedisplayskip}{\equspace}
	\setlength{\belowdisplayskip}{\equspace}
    W_{2}^{2}(\mathcal{N}_a, \mathcal{N}_b) = \left\|\left(\left[c x_{a}, c y_{a}, \frac{w_{a}}{2}, \frac{h_{a}}{2}\right]^{\mathrm{T}},\left[c x_{b}, c y_{b}, \frac{w_{b}}{2}. \frac{h_{b}}{2}\right]^{\mathrm{T}}\right)\right\|_{2}^{2}.
\end{equation}

However, $W_{2}^{2}(\mathcal{N}_a, \mathcal{N}_b)$ is a distance metric, and cannot be directly used as similarity metric (\ie, a value between 0 and 1 as IoU). Therefore, we use its exponential form normalization and obtain the new metric dubbed Normalized Wasserstein Distance (NWD): 
\begin{equation}
    \setlength{\abovedisplayskip}{\equspace}
	\setlength{\belowdisplayskip}{\equspace}
    NWD\left(\mathcal{N}_a, \mathcal{N}_b\right)=\exp \left(-\frac{\sqrt{W_{2}^2\left(\mathcal{N}_a, \mathcal{N}_b\right)}}{C}\right),
\end{equation}

where $C$ is a constant closely related to the dataset. In the following experiments, we empirically set $C$ to the average absolute size of AI-TOD and achieve the best performance. Moreover, we observe that $C$ is robust in a certain range, details will be shown in supplementary materials.

 \begin{figure}[t]
    \centering
    \subfigure
    {
        \label{fig:offset1}
        \includegraphics[width=0.22\linewidth]{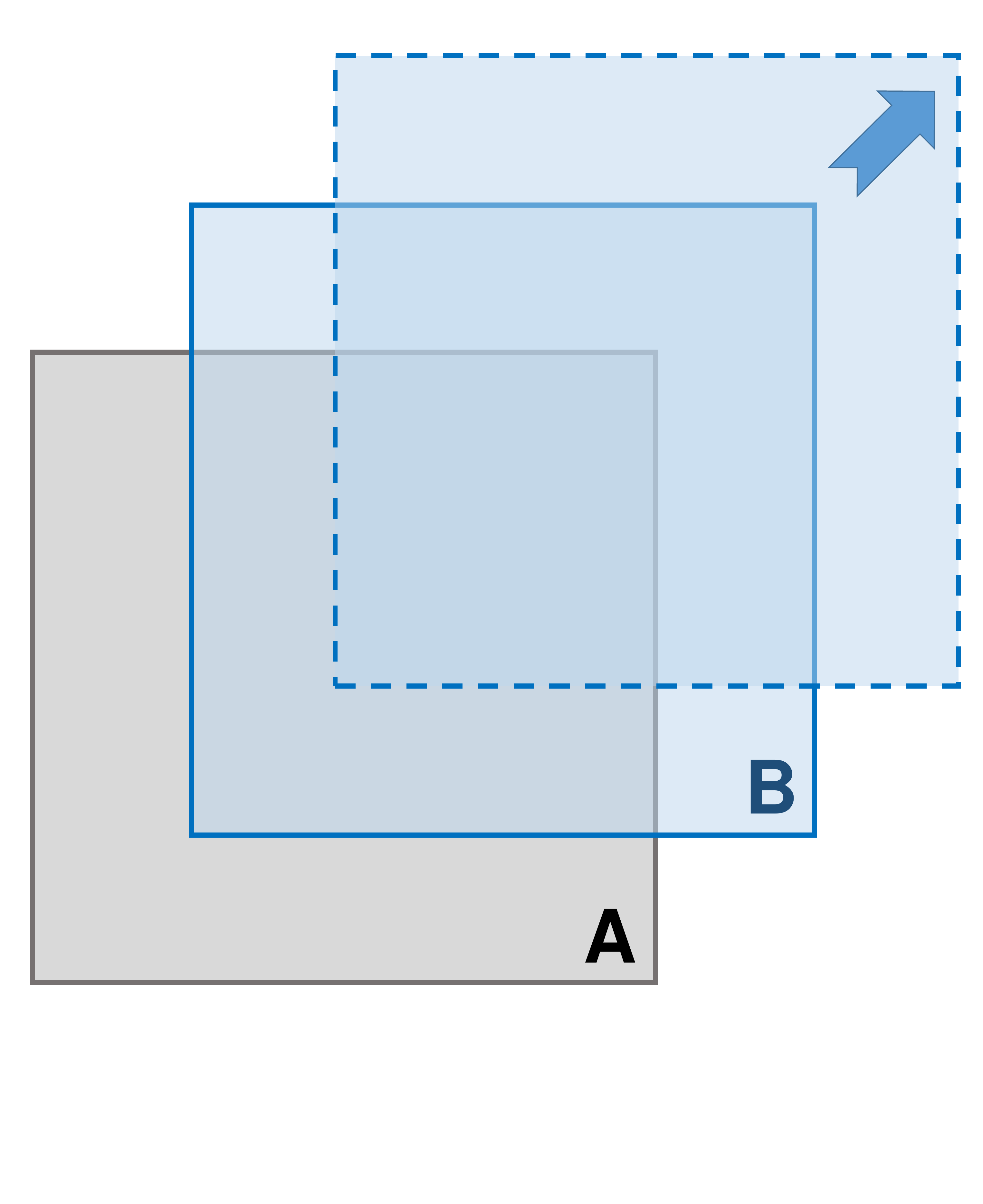}
    }
    \subfigure
    {
        \label{fig:iou1}
        \includegraphics[width=0.36\linewidth]{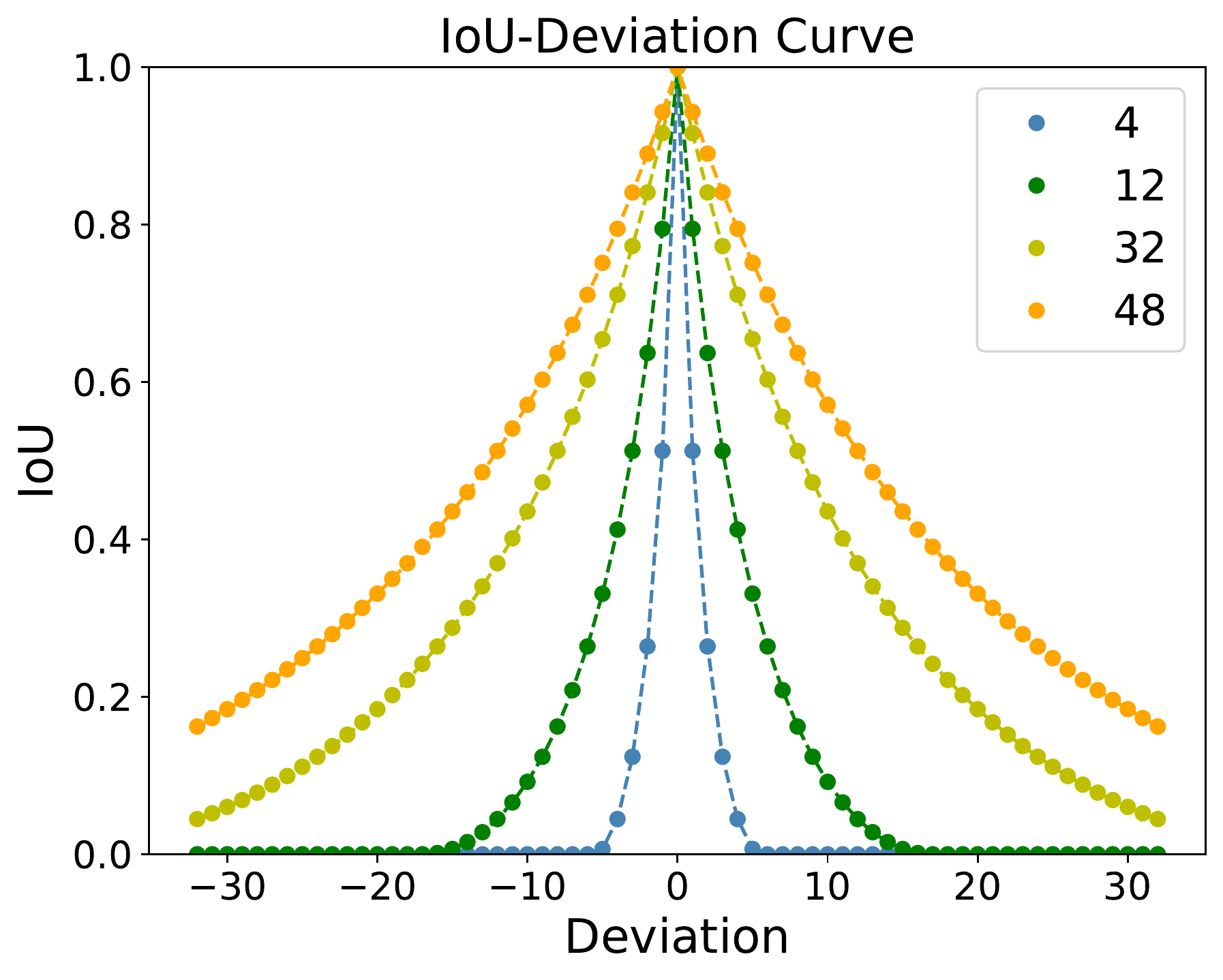}
    }
    \subfigure
    {
        \label{fig:nwd1}
        \includegraphics[width=0.36\linewidth]{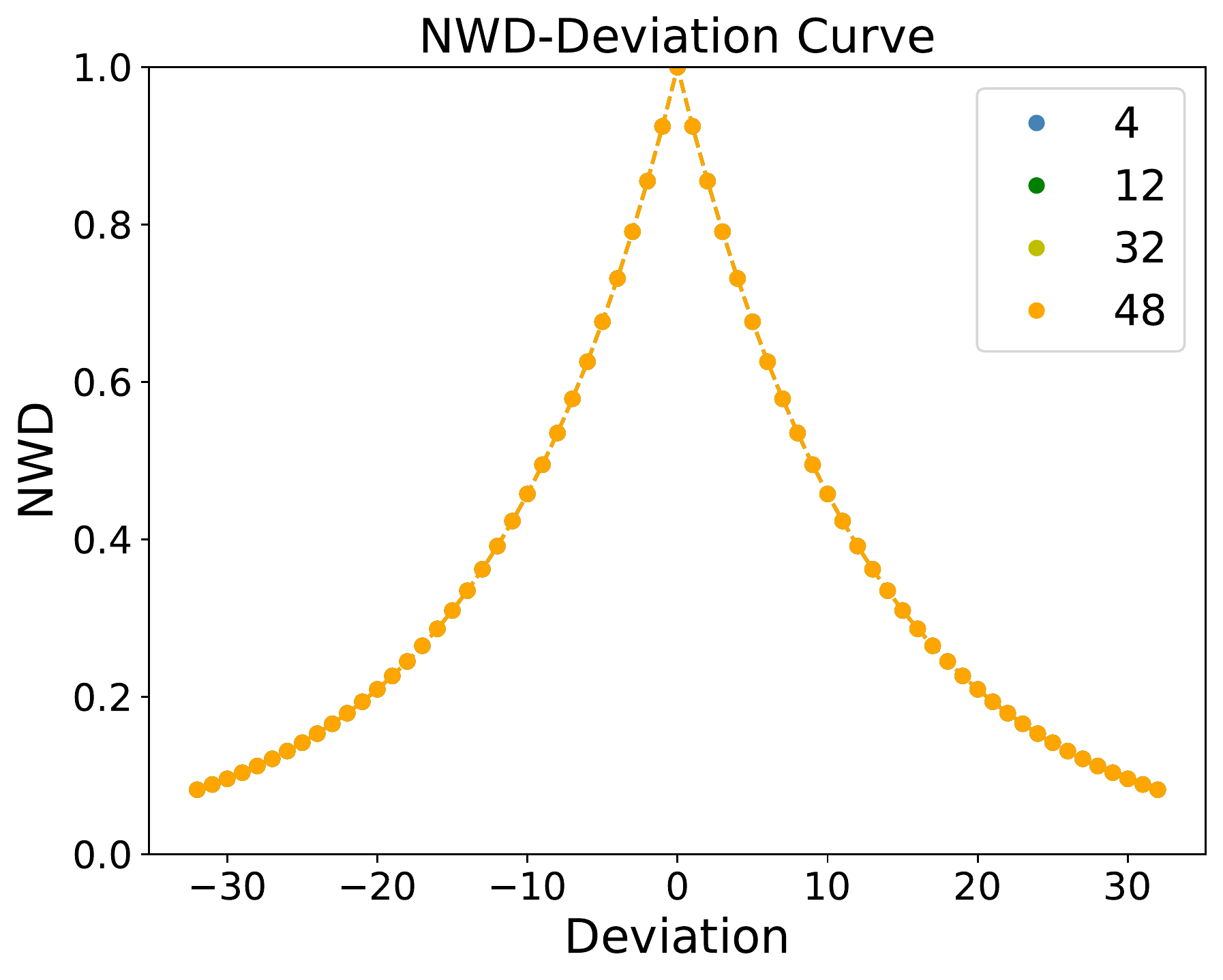}
    }
    \subfigure
    {
        \label{fig:offset2}
        \includegraphics[width=0.22\linewidth]{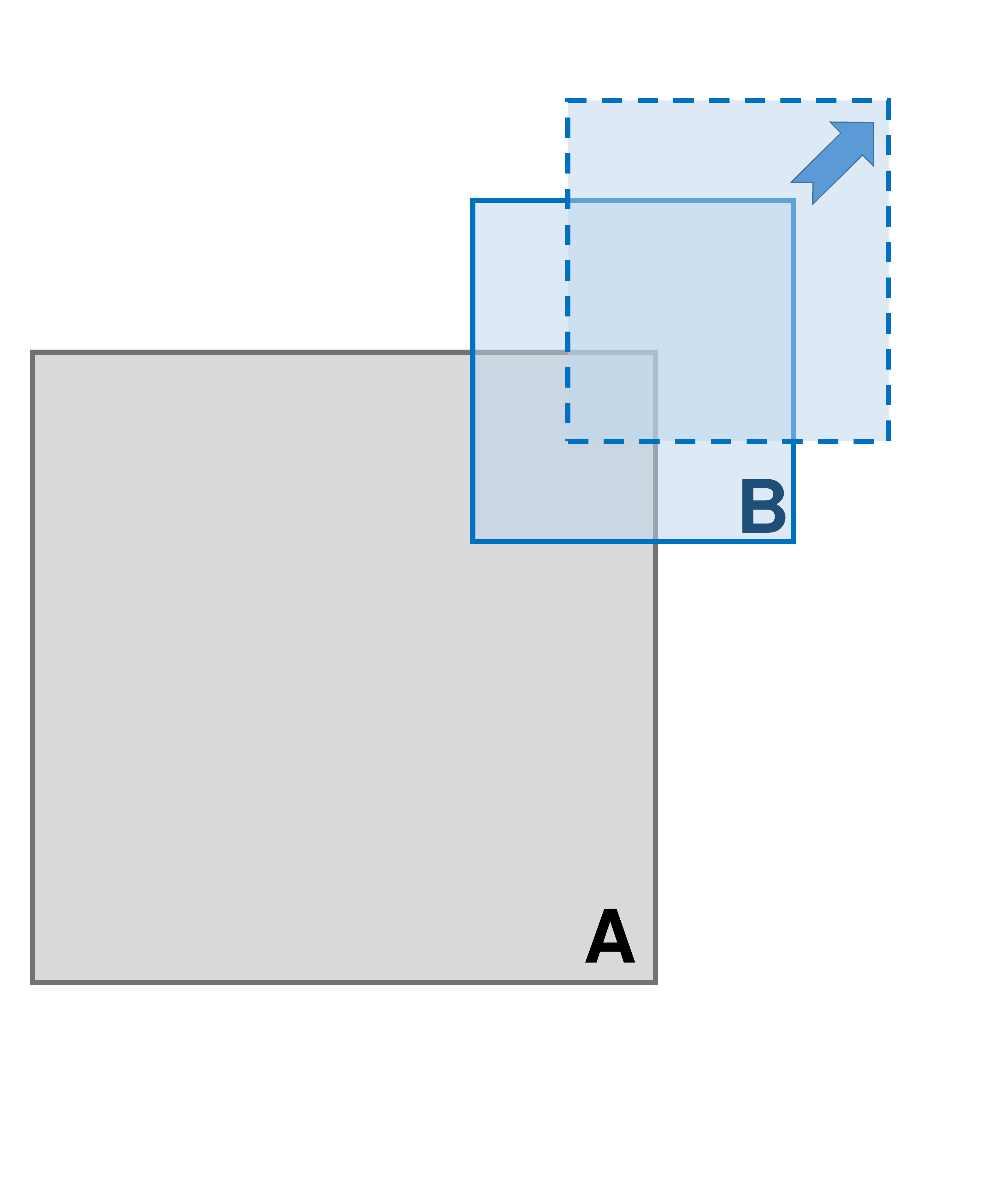}
    }
    \subfigure
    {
        \label{fig:iou2}
        \includegraphics[width=0.36\linewidth]{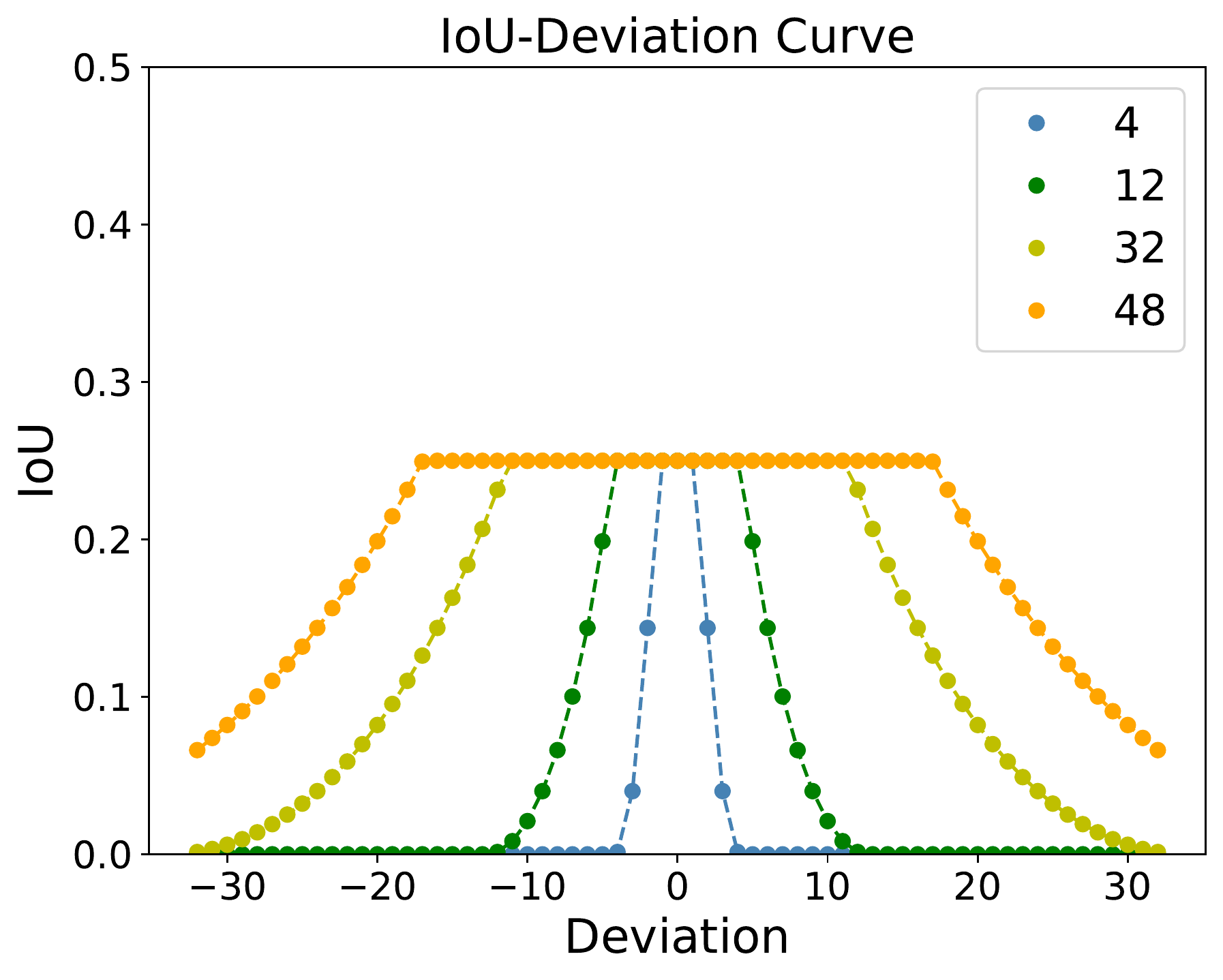}
    }
    \subfigure
    {
        \label{fig:nwd2}
        \includegraphics[width=0.36\linewidth]{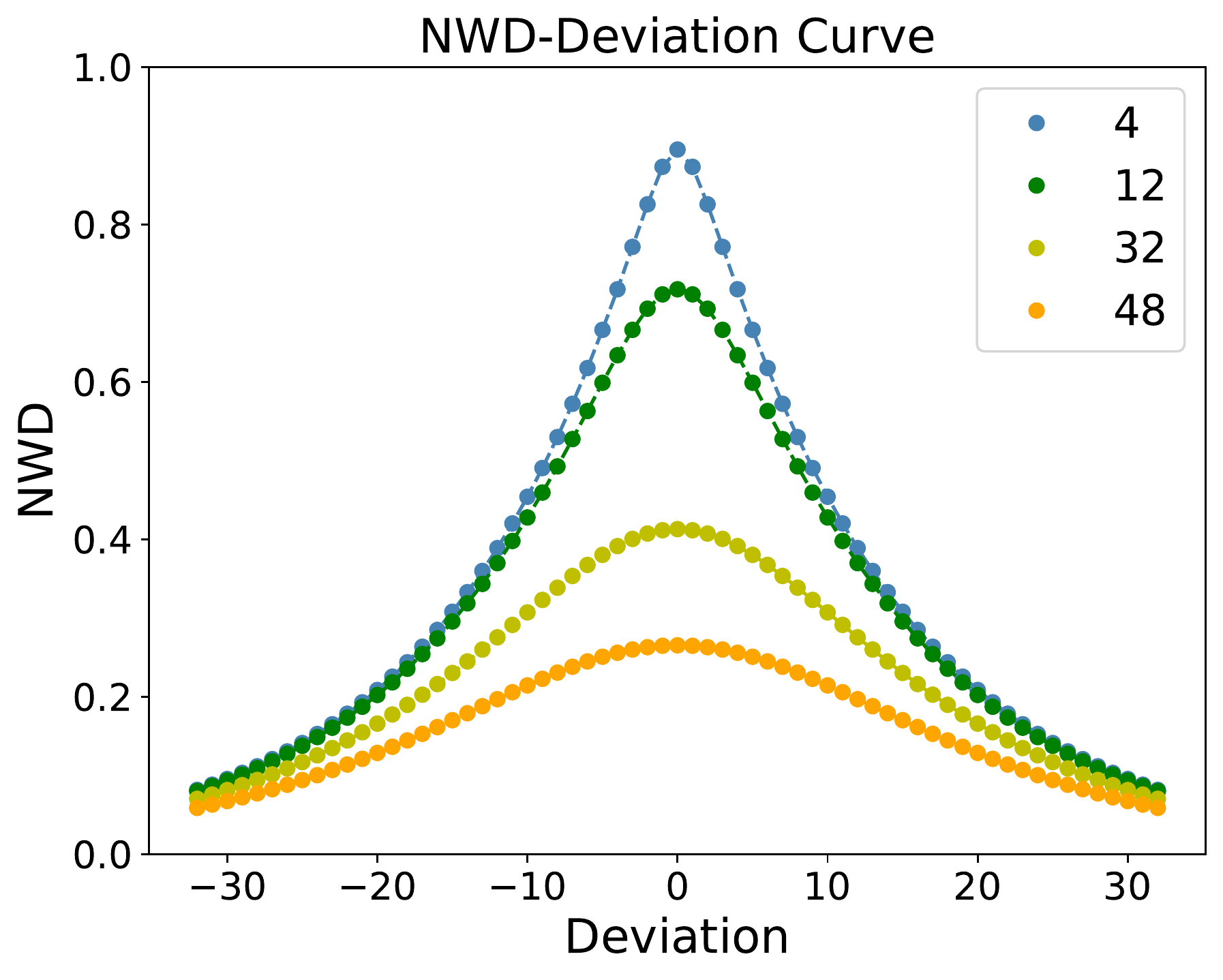}
    }
    \vspace{\fixedvskip}
    \caption{A comparison between IoU-Deviation Curve and NWD-Deviation Curve in two different scenarios. The abscissa value denotes the number of pixels deviation between the center points of $A$ and $B$, the ordinate value denotes the corresponding metric value. Note that the location of bounding box can only change discretely, the Value-Deviation curve is presented in the form of scatter diagram. }
    \label{fig::deviation_analysis}
    \vspace{\fixedvskip}
\end{figure}

Compared with IoU, NWD has the following advantages for detecting tiny objects: (1) scale invariance, (2) smoothness to location deviation, (3) the capability of measuring the similarity between non-overlapping or mutually inclusive bounding boxes. As shown in Fig.~\ref{fig::deviation_analysis}, without losing generality, we discuss the change of metric value in the following two scenarios. In the first row of Fig.~\ref{fig::deviation_analysis}, we keep box $A$ and $B$ the same scale and move away $B$ along the diagonal of $A$. It can be seen that the four curves of NWD completely coincide, which indicates that NWD is insensitive to the scale variance of boxes. Moreover, we can observe that IoU is too sensitive to minor location deviation, but the NWD change resulting from location deviation is more smooth. The smoothness to location deviation indicates a possibility of a better distinction between \posneg samples than IoU under the same threshold. In the second row of Fig.~\ref{fig::deviation_analysis}, we set the side length of $B$ to half of the side length of $A$ and move away $B$ along the diagonal of $A$. Compared with IoU, the curve of NWD is much more smooth and it can consistently reflect the similarity between $A$ and $B$ even if $|A\cap B| = A~{\rm or}~B$ and $|A\cap B| = 0$.

\subsection{NWD-based Detectors}

The proposed NWD can be easily integrated into any anchor-based detector to replace IoU. Without loss of generality, the representative anchor-based Faster R-CNN is adopted to describe the usage of NWD. Specifically, all the modifications are conducted in the three parts which originally employ IoU, including \posneg label assignment, NMS and regression loss function. The details are as follows.

\textbf{NWD-based Label Assignment.} Faster R-CNN~\cite{Faster-R-CNN_2015_NIPS} consists of two networks: RPN for generating region proposals and R-CNN~\cite{Fast-R-CNN_2015_ICCV} for detecting objects based on these proposals. The RPN and R-CNN both include label assignment process. For the RPN, anchors of different scales and ratios are firstly generated, and then binary labels are assigned to the anchors for training the classification and regression head. For the R-CNN, the label assignment process is similar with the RPN, and the difference is that the input of R-CNN is the output of RPN. In order to overcome the aforementioned shortcomings of IoU in tiny object detection, we design NWD-based label assignment strategy, which utilizes NWD to assign labels. Specifically, for training RPN, the positive label will be assigned to two kinds of anchors: (1) the anchor with the highest NWD value with a \gt box and the NWD value is larger than $\theta_{n}$ or (2) the anchor that has the NWD value higher than the positive threshold $\theta_{p}$ with any \gt. Accordingly, the negative label will be assigned to the anchor if its NWD value is lower than the negative threshold $\theta_{n}$ with all \gt boxes. In addition, the anchors that are neither assigned positive labels nor negative labels do not participate in the training process. Note that, in order to apply NWD to anchor-based detectors directly, $\theta_{p}$ and $\theta_{n}$ as the original detectors are used in the experiments.

\textbf{NWD-based NMS.} NMS is an integral part of the object detection pipeline to suppress the redundant prediction bounding boxes, in which the IoU metric is applied. First, it sorts all prediction boxes based on their scores. The prediction box $\mathcal{M}$ with the highest score is selected and all other prediction boxes with a significant overlap (using a pre-defined threshold $N_{t}$) with $\mathcal{M}$ are suppressed. This process is recursively applied on the remaining boxes. However, the sensitivity of IoU to tiny object will make the IoU values lower than $N_{t}$ for lots of prediction boxes, which further leads to false positive predictions. To handle this problem, we suggest that NWD is a better criterion for NMS in tiny object detection since NWD overcomes the scale sensitivity problem. Moreover, the NWD-based NMS is flexible to be integrated into any tiny object detector with only a few codes.

\textbf{NWD-based Regression Loss.} IoU-Loss~\cite{Unitbox_2016_ACMM} is introduced to eliminate the performance gap between training and testing~\cite{GIoU_loss_2019_CVPR}. However, IoU-Loss cannot provide gradient for optimizing network in the following two cases: (1) there is no overlap between the predicted bounding box $P$ and the ground-truth box $G$ (\ie, $|P\cap G| = 0$) or (2) box $P$ contains box $G$ completely or vice versa (\ie, $|P\cap G| = P~{\rm or}~G$). In addition, these two cases are very common for tiny objects. Specifically, on one hand, the deviation of a few pixels in $P$ will cause no overlap between $P$ and $G$, on the other hand, the tiny object is easy to be false predicted, leading to $|P\cap G| = P~{\rm or}~G$. Therefore, IoU-Loss is not suitable for tiny object detector. Although CIoU and DIoU can handle above two situations, they are sensitive to the location deviation of the tiny objects since they are both based on IoU. To handle above problems, we design the NWD metric as loss function by:
\begin{equation}
    \setlength{\abovedisplayskip}{\equspace}
	\setlength{\belowdisplayskip}{\equspace}
   \mathcal{L}_{NWD} = 1 - NWD\left(\mathcal{N}_p, \mathcal{N}_g\right),
\end{equation}
where $\mathcal{N}_{p}$ is the Gaussian distribution model of prediction box $P$, $\mathcal{N}_{g}$ is the Gaussian distribution model of \gt box $G$. According to the introduction in Sec.~\ref{sec::nwd_definition}, NWD-based loss can provide gradient even in both cases $|P\cap G| = 0$ and $|P\cap G| = P~{\rm or}~G$.

\section{Experiments}

\label{experiment_details}
We evaluate the proposed method on AI-TOD~\cite{AI-TOD_2020_ICPR} and VisDrone2019~\cite{visdrone2019_2019_iccvw} datasets. The ablation study is conducted on AI-TOD, which is a challenging dataset designed for tiny object detection. It comes with eight categories, $700,621$ object instances across $28,036$ aerial images with $800\times 800$ pixels. The mean absolute size of AI-TOD is only $12.8$ pixels, which is much smaller than other object detection dataset like PASCAL VOC ($156.6$ pixels)~\cite{PASCAL-VOC_2015-IJCV}, MS COCO ($99.5$ pixels)~\cite{COCO_2014_ECCV}, and DOTA ($55.3$ pixels)~\cite{DOTA_2018_CVPR}. In addition, VisDrone2019~\cite{visdrone2019_2019_iccvw} is an UAV dataset for object detection. It consists of 10,209 images with 10 categories. VisDrone2019 has many complex scenes and large numbers of tiny objects since images are captured in different places at different height.

We adopt the same evaluation metric as AI-TOD~\cite{AI-TOD_2020_ICPR} dataset, including $\rm{AP}$, $\rm{AP_{0.5}}$, $\rm{AP_{0.75}}$, $\rm{AP}_{vt}$, $\rm{AP}_{t}$, $\rm{AP}_{s}$ and $\rm{AP}_{m}$. Specifically, $\rm{AP}$ is averaged mAP across different IoU thresholds IoU=$\{0.5, 0.55, \cdots, 0.95\}$, $\rm{AP_{0.5}}$ and $\rm{AP_{0.75}}$ are APs at IoU threshold of $0.5$ and $0.75$, respectively. In addition, $\rm{AP}_{vt}$, $\rm{AP}_{t}$, $\rm{AP}_{s}$ and $\rm{AP}_{m}$ are for \textit{very tiny} ($2$-$8$ pixels), \textit{tiny} ($8$-$16$ pixels), \textit{small} ($16$-$32$ pixels) and \textit{medium} ($32$-$64$ pixels) scale evaluation in AI-TOD~\cite{AI-TOD_2020_ICPR}.

We conduct all the experiments on a computer with 4 NVIDIA Titan X GPUs, and the codes are used for our experiments are based on MMdetection~\cite{mmdetection_2019_arXiv} code library. The ImageNet~\cite{ImageNet_2015_IJCV}
pretrained ResNet-50~\cite{ResNet_2016_CVPR} with FPN~\cite{FPN_2017_CVPR} is used as the backbone, unless specified otherwise. All models are trained using the SGD optimizer for 12 epochs with 0.9 momentum, 0.0001 weight decay and 8 batch size. We set the initial learning rate as 0.01 and decay it at epoch 8 and 11 by a factor of 0.1. Besides, the batch size of RPN and Fast R-CNN are set to 256 and 512, respectively, and the sampling ratio of positive and negative samples is set to 1/3. The number of proposals generated by RPN is set to 3000. In the inference stage, we use the preset score 0.05 to filter out background bounding boxes, and NMS is applied with the IoU threshold 0.5. The above training and inference parameters are used in all experiments, unless specified otherwise.

\begin{table*}[t]
	\centering
	\setlength{\belowcaptionskip}{0.2cm}
	\caption{Comparison of different metrics in label assignment, NMS and loss function.}
	\begin{tabular}{|c|ccc|ccc|ccc|}  
	\hline
	\multirow{2}{*}{Metric}  & \multicolumn{3}{c|}{Assigning}  & \multicolumn{3}{c|}{NMS}  & \multicolumn{3}{c|}{Loss}  \\
	 &  $\rm{AP}$ & $\rm{AP_{0.5}}$ & $\rm{AP_{t}}$& $\rm{AP}$ & $\rm{AP_{0.5}}$ & $\rm{AP_{t}}$ &  $\rm{AP}$ & $\rm{AP_{0.5}}$ & $\rm{AP_{t}}$\\
	\hline
	DIoU  & 5.4 & 11.3 & 3.9  & 11.2 & 26.8 & 7.8 & 10.7 & 25.1 & 6.7 \\
	CIoU & 5.9 & 12.5 & 4.4 & 10.9 & 25.7 & 7.2 & 10.6 & 24.9 & 6.8  \\
	GIoU & 11.0 & 26.5  & 7.7 & 11.5 & 26.5 & 7.6 & 10.9 & 25.1 & 6.9  \\
	IoU & 11.1 & 26.5  & 7.8 & 11.1 & 26.5 & 7.8 & 10.8 & 25.3 & 7.1 \\
	NWD & \textbf{16.1} & \textbf{43.8}  & \textbf{17.4} & \textbf{11.9} & \textbf{27.5}  & \textbf{8.0} & \textbf{12.1} & \textbf{27.5}  & \textbf{8.9} \\
	\hline
	\end{tabular}
	\label{tab:metrics}
	\vspace{\fixedvskip}
\end{table*}

\subsection{Comparison with Other Metrics based IoU}

There are some IoU-based metrics can be used to measure the similarity between bounding boxes as mentioned in Sec.~\ref{sec::related_work}. In this work, we re-implement the aforementioned four metrics (\ie IoU, GIoU, CIoU and DIoU) and our proposed NWD on the same basic network (\ie Faster R-CNN) to compare their performance on tiny objects. Specifically, they are applied in label assignment, NMS and loss function, respectively. Experimental results on AI-TOD dataset are shown in Tab.~\ref{tab:metrics}.

\textbf{Comparison in label assignment.} Note that the metric in assigning modules of RPN and R-CNN are both modified. It can be seen that NWD achieves the highest AP of $16.1\%$ and improves $9.6\%$ on $\rm{AP_{t}}$ when comparing with IoU metric, revealing that the NWD-based label assignment can provide more high quality training samples for tiny objects. In addition, to analyze the essential of the improvement, We make a group of statistical experiment. Specifically, we respectively calculate the average number of positive anchors matched by each \gt box when using IoU, GIoU, DIoU, CIoU and NWD under the same default threshold, the number is 0.72, 0.71, 0.19, 0.19 and 1.05 respectively. It can be found that only NWD can ensure a considerable number of positive training samples. Moreover, although simply lowering the threshold of IoU-based metrics can provide more positive anchors for training, the performance of IoU-based tiny object detector after threshold fine-tuning is not better than the performance of NWD-based detector, which will be further discussed in supplementary materials. It attributes to the fact that NWD can solve the sensitivity of IoU to tiny object location deviation.

\textbf{Comparison in NMS.} We only modify the NMS module of RPN in this experiment since only the NMS in RPN can directly affect the training processing of detector. It can be seen that using different metrics to filter out redundant predictions during training can also affect the detection performance. Concretely, NWD achieves the best AP of $11.9\%$, which is $0.8\%$ higher than the commonly used IoU. This implies that the NWD is a better metric for filtering out redundant bounding boxes when detecting tiny objects.

\textbf{Comparison in loss function.} Note that we modify the loss function both in RPN and R-CNN, which can both affect the convergence of the detector. It can also be seen that NWD-based loss function achieves the highest AP of $12.1\%$.

\subsection{Ablation Study}
 In this section, Faster R-CNN~\cite{Faster-R-CNN_2015_NIPS} are used as the baseline, and it consists of two stages: RPN and R-CNN. Our proposed method can both be applied in the label assignment, NMS, loss function module of RPN and R-CNN, therefore there are totally six modules that can be switched from IoU metric to NWD metric. To verify the effectiveness of our proposed method in different modules, we make the following two groups of ablation study: comparison of applying NWD into one of the six modules and comparison of applying NWD into all modules in RPN or R-CNN.

\begin{table*}[t]
	\centering
	\setlength{\belowcaptionskip}{0.2cm}
	\caption{Ablation experiments when NWD is applied to single module.}
	\begin{tabular}{|c|cc|cc|cc|c|}  
	\hline
	\multirow{2}{*}{Method}  & \multicolumn{2}{c|}{Assigning}  & \multicolumn{2}{c|}{NMS}  & \multicolumn{2}{c|}{Loss} & \multirow{2}{*}{AP} \\
	&  RPN & R-CNN & RPN & R-CNN & RPN & R-CNN & \\
	\hline
	Baseline & & & & & & & 11.1 \\
	\hline
	\multirow{6}{*}{NWD}  & \checkmark &  &  &  &  &  & \textbf{17.3} \\
	                      &  & \checkmark &  &  &  &  & 14.3 \\
	                      &  &  & \checkmark &  &  &  & 11.9 \\
	                      &  &  &  & \checkmark &  &  & 10.8 \\
	                      &  &  &  &  & \checkmark  &  & 12.1 \\
	                      &  &  &  &  &  & \checkmark & 12.4 \\
	\hline
	\end{tabular}
	\label{tab:single}
\end{table*}

\textbf{Applying NWD into single module.} Experimental results are shown in Tab.~\ref{tab:single}. Compared to baseline method, NWD-based assigning module in RPN and R-CNN respectively achieves the highest and second-highest AP improvement of 6.2\% and 3.2\%, which indicates that the problem of tiny object training label assignment resulting from IoU is the most noticeable, and our proposed NWD-based assignment strategy greatly improves assignment quality. It can also be observed that our proposed method improves the performance in 5 out of 6 modules, which significantly verifies the effectiveness of our NWD-based method. And the performance drop in NMS of R-CNN may owe to the fact that the default NMS threshold is sub-optimal, and it needs fine-tuning to boost the performance. 

\begin{table*}[t]
	\centering
    \setlength{\belowcaptionskip}{0.2cm}
	\caption{Ablation experiments when NWD is applied to multiple modules.}
	\begin{tabular}{|c|cc|cc|cc|c|c|}  
	\hline
	\multirow{2}{*}{Method} & \multicolumn{2}{c|}{Assigning}  & \multicolumn{2}{c|}{NMS}  & \multicolumn{2}{c|}{Loss} & AP & AP \\
      & RPN & R-CNN & RPN & R-CNN & RPN & R-CNN & 12 epochs & 24 epochs\\
	\hline
	Baseline & & & & & & & 11.1 & 12.6 \\
	\hline
	\multirow{3}{*}{NWD} & \checkmark &  & \checkmark &  &\checkmark  &  & \textbf{17.8} & \textbf{19.7} \\
	                     &  & \checkmark &  & \checkmark &  & \checkmark & 13.8 & 16.8 \\
	                     & \checkmark & \checkmark & \checkmark & \checkmark & \checkmark & \checkmark & 15.2 & 18.8 \\
	                         
	\hline
	\end{tabular}
	\label{tab:multi1}
	\vspace{\fixedvskip}
\end{table*}

\textbf{Applying NWD into multiple modules.} Tab.~\ref{tab:multi1} lists the experimental results. When training for 12 epochs, the detection performance all achieves significant improvement when using NWD in RPN, R-CNN or all modules. And the best performance of 17.8\% is achieved when we apply NWD into all three modules of RPN. However, we find that when using NWD in all six modules, the AP has a drop of 2.6\% compared with merely using NWD in RPN. In order to analyze the reason for performance drop, we add a group of experiments and train the network for 24 epochs. It can be seen that the performance gap decreases from 2.6\% to 0.9\%, which reveals that the network needs more time to converge when using NWD in R-CNN. Therefore, we only use NWD in RPN to achieve a considerable performance improvement with less time in the following experiments.

\begin{figure*}[h]
    \centering
    \subfigure{
    \includegraphics[width=0.232\linewidth]{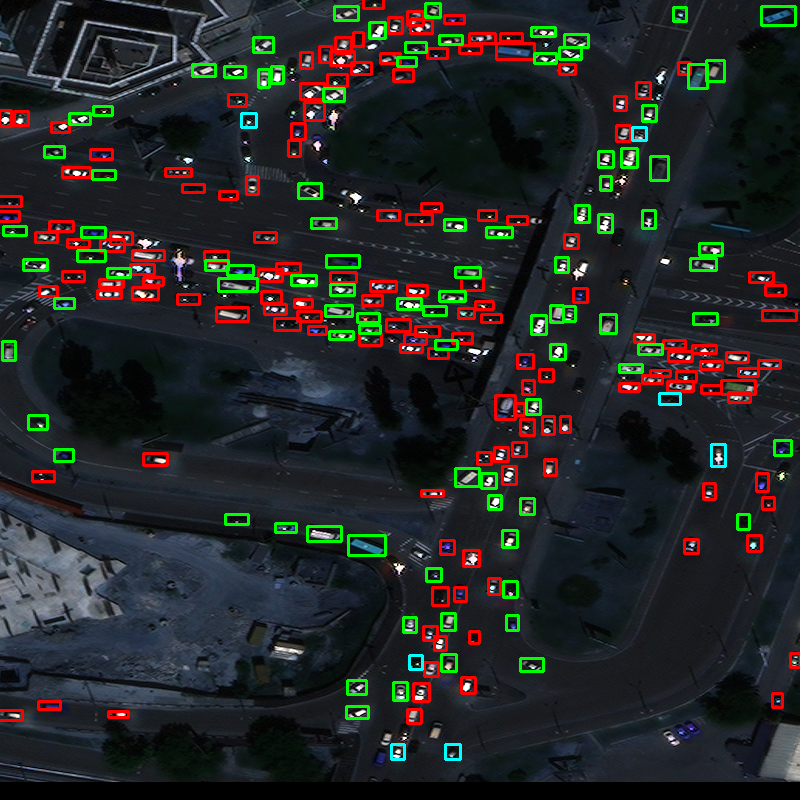}
    \label{fig4a1}
    }
    \subfigure{
    \includegraphics[width=0.232\linewidth]{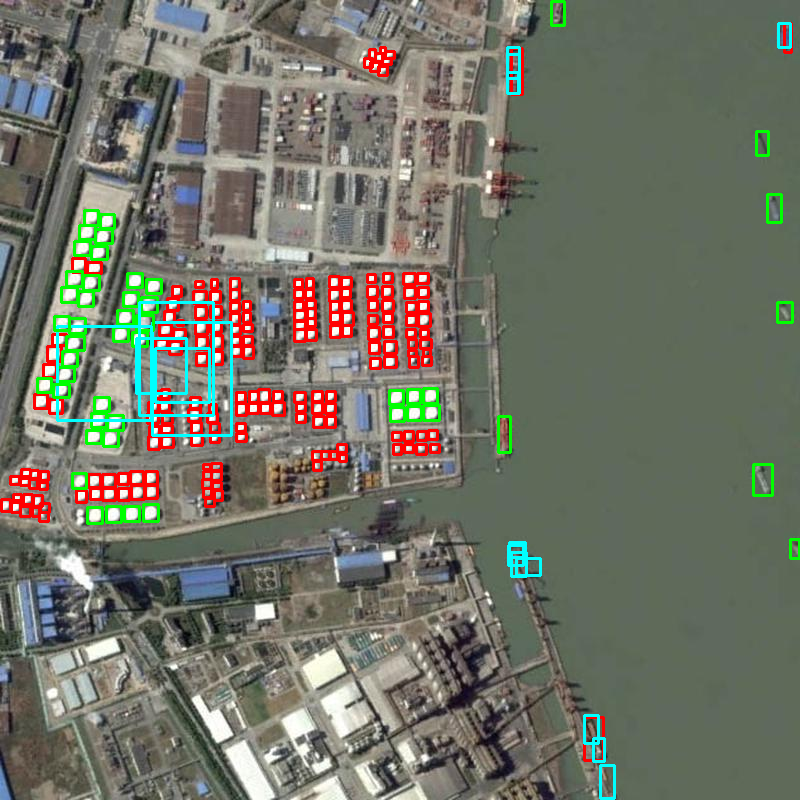}
    \label{fig4b1}
    }
    \subfigure{
    \includegraphics[width=0.232\linewidth]{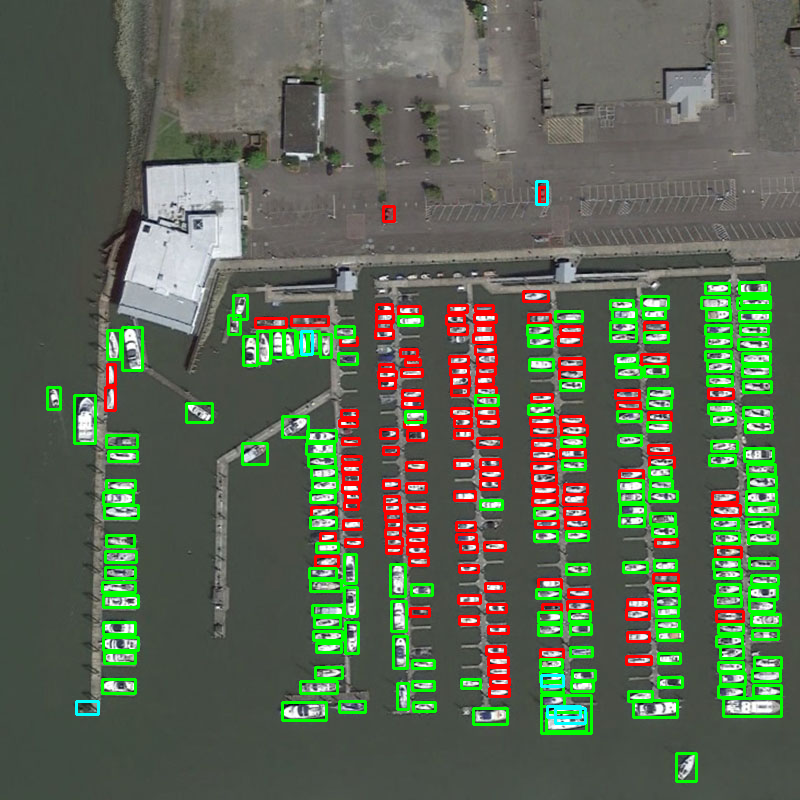}
    \label{fig4c1}
    }
    \subfigure{
    \includegraphics[width=0.232\linewidth]{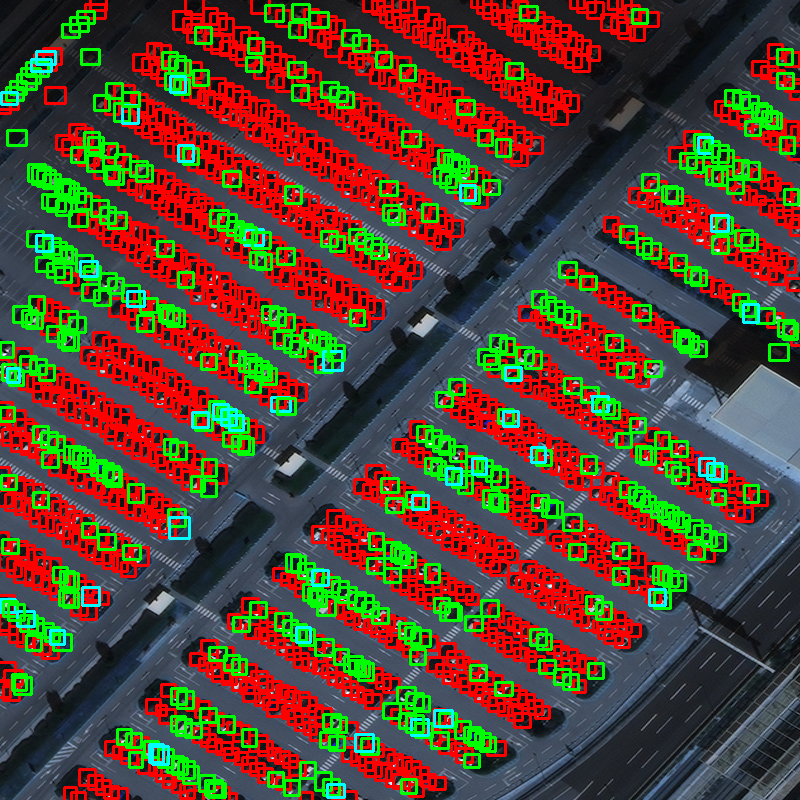}
    \label{fig4d1}
    }
    \\ \vspace{-0.2cm}
    \subfigure{
    \includegraphics[width=0.232\linewidth]{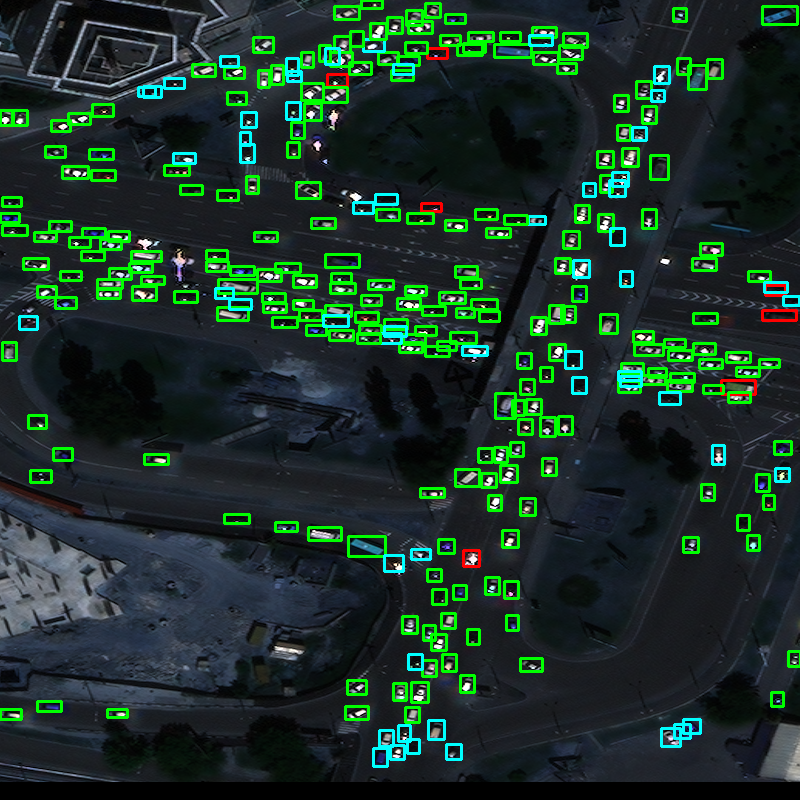}
    \label{fig4a2}
    }
    \subfigure{
    \includegraphics[width=0.232\linewidth]{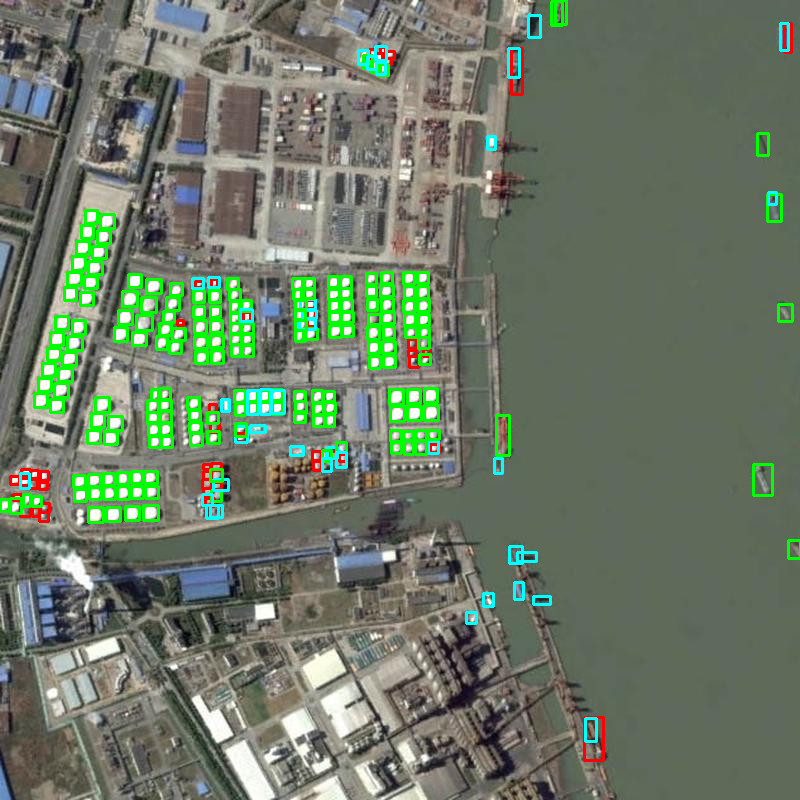}
    \label{fig4b2}
    }
    \subfigure{
    \includegraphics[width=0.232\linewidth]{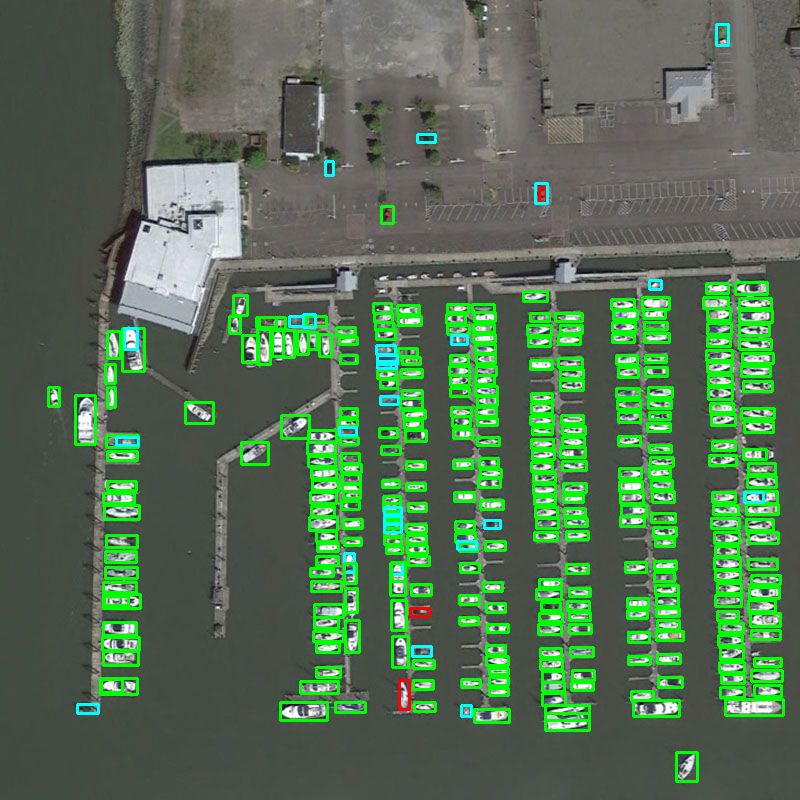}
    \label{fig4c2}
    }
    \subfigure{
    \includegraphics[width=0.232\linewidth]{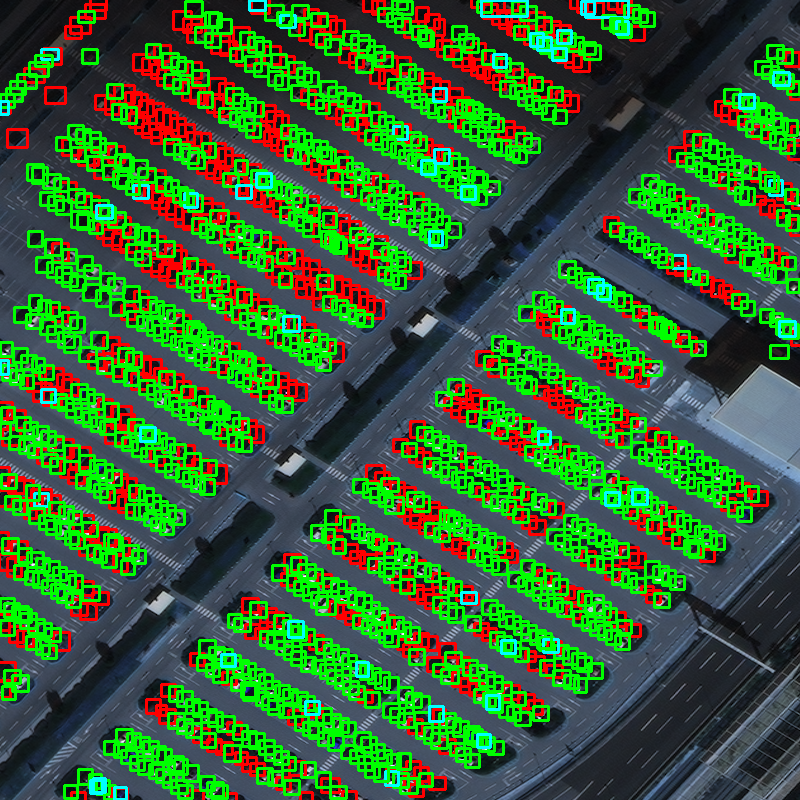}
    \label{fig4d2}
    }
    \vspace{\fixedvskip}
    \caption{Visualization of detection results using IoU-based detector (first row) and NWD-based detector (second row) of AI-TOD dataset. The green, blue and red boxes denote true positive (TP), false positive (FP) and false negative (FN) predictions, respectively.}
    \label{f5}
    \vspace{\fixedvskip}
\end{figure*}

\subsection{Main Results}

To reveal the effectiveness of NWD on TOD, we conduct experiments on tiny object detection datasets AI-TOD~\cite{AI-TOD_2020_ICPR} and VisDrone2019~\cite{visdrone2019_2019_iccvw}.

\textbf{Main results on AI-TOD.} To verify that NWD can be applied into any anchor-based detector and boost TOD performance, we select five baseline detectors, including one-stage anchor-based detectors (\ie, RetinaNet~\cite{Focal-Loss_2017_ICCV}, ATSS~\cite{atss_2020_cvpr}) and multi-stage anchor-based detectors (\ie, Faster R-CNN~\cite{Faster-R-CNN_2015_NIPS}, Cascade R-CNN~\cite{Cascade-R-CNN_2018_CVPR}, DetectoRS~\cite{DetectoRS_2020_arXiv}). Experimental results are shown in Tab.~\ref{tab:baselines}. It can be seen that $\rm{AP_{vt}}$ of current state-of-the-art detectors is extremely low and close to zero, that means they cannot produce satisfactory results on tiny objects. In addition, our proposed NWD-based detectors improve AP metric of RetinaNet, ATSS, Faster R-CNN, Cascade R-CNN and DetectoRS by 4.5\%, 0.7\%, 6.7\%, 4.9\% and 6.0\%, respectively. The performance improvement is even more obvious when objects are extremely tiny. It is worth noticing that NWD-based DetectoRS achieves state-of-the-art performance ($20.8\%$ AP) on AI-TOD.  Some visualization results using IoU-based detector (first row) and NWD-based detector (second row) on AI-TOD dataset are shown in Fig.~\ref{f5}. We can observe that NWD can significantly reduce false negative (FN) compared with IoU.

\textbf{Main results on Visdrone.} Besides AI-TOD, we use VisDrone2019~\cite{visdrone2019_2019_iccvw} which contains many tiny objects with different scenarios to verify the generalization of NWD-based detectors. The results are shown in Tab.~\ref{tab:visdrone}. It can be seen that NWD-based anchor-based detectors all achieve considerable improvements over their baselines.

\begin{table*}[t]
    \setlength{\belowcaptionskip}{0.1cm}
    \renewcommand{\arraystretch}{0.92}
    \caption{Quantitative comparison of the baselines and NWD (with *) on  AI-TOD {\tt test set}.}
	\centering
	\begin{tabular}{|l|c|ccccccc|}  
	\hline
	 Method                                     & Backbone & AP & $\rm{AP_{0.5}}$ & $\rm{AP_{0.75}}$ & $\rm{AP_{vt}}$ & $\rm{AP_{t}}$ & $\rm{AP_{s}}$ & $\rm{AP_{m}}$\\
	\hline
	 SSD-512~\cite{SSD_2016_ECCV}        	    & ResNet-50 	& 7.0	& 21.7  & 2.8  & 1.0 & 4.7 & 11.5 & 13.5  \\
	 TridentNet~\cite{Trident-Net_2019_ICCV}      & ResNet-50 	    & 7.5	& 20.9  & 3.6  & 1.0 & 5.8 & 12.6 & 14.0   \\
	 FoveaBox~\cite{FoveaBox_2020_TIP}          & ResNet-50 	& 8.1	& 19.8  & 5.1  & 0.9 & 5.8 & 13.4  & 15.9  \\
	 PepPonits~\cite{RepPoints_2019_ICCV}       & ResNet-50 	& 9.2	& 23.6  & 5.3  & 2.5 & 9.2 & 12.9  & 14.4  \\
	 FCOS~\cite{FCOS_2019_ICCV}                 & ResNet-50 	& 9.8	& 24.1  & 5.9  & 1.4 & 8.0 & 15.1  & 17.4  \\
	 CenterNet~\cite{CenterNet_2019_arXiv}      & DLA-34            & 13.4  & 39.2  & 5.0  & 3.8 & 12.1 & 17.7 & 18.9  \\
	 M-CenterNet~\cite{AI-TOD_2020_ICPR}        & DLA-34            & 14.5  & 40.7  & 6.4  & 6.1 & 15.0 & 19.4 & 20.4  \\
	\hline
	 RetinaNet~\cite{Focal-Loss_2017_ICCV}      & ResNet-50 	& 4.7	& 13.6  & 2.1  & 2.0 & 5.4 & 6.3  & 7.6   \\
	 RetinaNet*                                 & ResNet-50 	& 9.2	& 24.9  & 5.0  & 3.2 & 10.0 & 13.1  & 16.9   \\
	 \hline
	 ATSS~\cite{atss_2020_cvpr}                 & ResNet-50 	& 12.8	& 30.6  & 8.5  & 1.9 & 11.6 & 19.5 & 29.2   \\
	 ATSS*                                      & ResNet-50 	& 13.5	& 33.2  & 8.6  & 2.1 & 11.1 & 20.9 & 31.9   \\
	 \hline
	 Faster R-CNN~\cite{Faster-R-CNN_2015_NIPS} & ResNet-50 	& 11.1	& 26.3  & 7.6  & 0.0 & 7.2 & 23.3 & 33.6   \\
	 Faster R-CNN*                              & ResNet-50 	& 17.8	& 43.8  & 11.0  & 2.5 & 17.0 & 26.1 & 34.3 \\
	 \hline
	 Cascade R-CNN~\cite{Cascade-R-CNN_2018_CVPR}& ResNet-50 	& 13.8	& 30.8  & 10.5 & 0.0 & 10.6 & 25.5 &  36.6 \\
	 Cascade R-CNN*                             & ResNet-50 	& 18.7	& 44.2  & 12.9 & 3.6 & 17.4 & 26.5 & 35.6 \\
	 \hline
	 DetectoRS~\cite{DetectoRS_2020_arXiv}      & ResNet-50 	& 14.8	& 32.8  & 11.4 & 0.0 & 10.8 & 28.3 & 38.0 \\
	 DetectoRS*                                 & ResNet-50 	& \textbf{20.8}	& \textbf{49.3}  & \textbf{14.3} & \textbf{6.4} & \textbf{19.7} & \textbf{29.6} & \textbf{38.3} \\
	\hline
	\end{tabular}
	\label{tab:baselines}
	\vspace{-2mm}
\end{table*}

\begin{table}
	\centering
	\setlength{\belowcaptionskip}{0.1cm}
	\renewcommand{\arraystretch}{0.92}
	\caption{Quantitative comparison of the baselines and NWD (with *) on VisDrone2019 {\tt val set}.}
	\begin{tabular}{|l|cc|cc|}  
	\hline
	Method  & Faster R-CNN & Faster R-CNN* & Cascade R-CNN & Cascade R-CNN*  \\
	\hline
	$\rm{AP_{0.5}}$  & 38.0 & \textbf{38.5} & 38.5 & \textbf{40.3} \\
	$\rm{AP_{vt}}$   & 0.1 & \textbf{3.8} & 0.5 & \textbf{2.9} \\
	$\rm{AP_{t}}$   & 6.2 & \textbf{10.2} & 6.8 & \textbf{11.1} \\
	$\rm{AP_{s}}$   & 20.0 & \textbf{21.4} & 21.4 & \textbf{22.2} \\
	\hline
	\end{tabular}
	\label{tab:visdrone}
	\vspace{-5mm}
\end{table}

\section{Conclusion}
In this paper, we observe that IoU-based metrics is sensitive to the location deviation of tiny objects, which drastically deteriorates the tiny object detection performance. To handle this problem, we propose a new metric dubbed Normalized Wasserstein Distance (NWD) to measure the similarity between bounding boxes for tiny objects. Based on that, we further present a novel NWD-based tiny object detector by embedding NWD into label assignment, non-maximum suppression, and loss function of anchor-based detectors to replace original IoU metric. Experimental results show that our proposed method can improve the tiny object detection performance by a large margin and achieve state-of-the-art on AI-TOD dataset.

\medskip

{
\bibliographystyle{ieee_fullname}
\bibliography{chang, Jinwang-Papers}
}

\end{document}